\documentclass{article}

    \PassOptionsToPackage{numbers, compress}{natbib}
 \usepackage[preprint]{neurips_2026}

\usepackage[utf8]{inputenc} % allow utf-8 input
\usepackage[T1]{fontenc}    % use 8-bit T1 fonts
\usepackage{url}            % simple URL typesetting
\usepackage{booktabs}       % professional-quality tables
\usepackage{amsfonts}       % blackboard math symbols
\usepackage{nicefrac}       % compact symbols for 1/2, etc.
\usepackage{microtype}      % microtypography
\usepackage[table, dvipsnames, svgnames, x11names]{xcolor}         % colors

\usepackage[colorlinks,
            linkcolor=red,
            anchorcolor=green,
            citecolor=NavyBlue
            ]{hyperref}
\usepackage{enumitem}
\usepackage{multirow}
\newcommand{\token}[1]{\texttt{\small\textless#1\textgreater}}
\usepackage{arydshln} % 导入虚线宏包
\usepackage{subfig}
\usepackage{tcolorbox}
\usepackage{soul, color}

\newcommand{\cf}{\emph{cf. }}

\title{Scaling Retrieval-Augmented Reasoning with Parallel Search and Explicit Merging}
\author{
\\
 \textbf{Jiabei Liu\textsuperscript{1*}},
 \textbf{Wenyu Mao\textsuperscript{1*}},
 \textbf{Junfei Tan\textsuperscript{1}},
 \textbf{Chunxu Shen\textsuperscript{2$\dag$}},
 \textbf{Lingling Yi\textsuperscript{2}},
\\
 \textbf{Jiancan Wu\textsuperscript{1$\dag$}},
 \textbf{Xiang Wang\textsuperscript{1$\dag$}},
\\
[3mm]
$^1$ University of Science and Technology of China \\
$^2$ Independent Researcher \\
$^*$ Equal contribution. $^\dagger$ Corresponding author.
}

\usepackage{graphicx}
\usepackage{amsmath}

\begin{document}

\maketitle

\begin{abstract}
% Background
%% Advantage
% Deep search agents have shown promising capabilities in reasoning and information retrieval. 
% % Deep search agents have demonstrated strong capabilities in multi-step retrieval-augmented reasoning, enabling large language models to access external knowledge for complex tasks.
% %% Problem
% However, existing approaches often rely on single-route retrieval paradigm, which may underutilize each search step and introduce noisy information, limiting the efficiency and quality of reasoning.
% % Our Method
% %% What
% In this work, we propose MultiSearch, an agentic reinforcement learning framework designed to enhance the retrieval-during-reasoning ability of large language models. 
% %% How - First
% MultiSearch introduces a multi-route retrieval paradigm combined with an explicit merging step, encouraging the agent to collect more extensive and higher-quality evidence. 
% %% How - Moreover
% The agent is trained using group reward–decoupled normalization policy optimization, leveraging both outcome-based reward and auxiliary step-specific rewards. 
% % Exp
% Experiments on seven benchmarks show that MultiSearch achieves an improvement of $x\%$ in average accuracy over baseline methods, validating its potential to enhance search and reasoning capabilities in question-answering tasks.
% Analysis
% Further analysis suggests that 

Deep search agents have proven effective in enhancing LLMs by retrieving external knowledge during multi-step reasoning. However, existing methods often generate a single query for retrieval at each reasoning step, limiting information coverage and introducing high noise. This may result in low signal-to-noise ratios (SNR) during search, degrading reasoning accuracy and leading to unnecessary reasoning steps. In this paper, we introduce MultiSearch, an RL-based framework that addresses these limitations through multi-query retrieval and explicit merging of retrieved information. At each reasoning step, MultiSearch generates queries from multiple perspectives and retrieves external information in parallel, expanding the scope of relevant information and mitigating the reliance on any single retrieval result. Then, the agent consolidates and refines retrieved information at the merging process, improving the SNR and ensuring more accurate reasoning. Additionally, we propose a reinforcement learning framework with a multi-process reward design to optimize agents for both multi-query retrieval and information consolidation.
Extensive experiments on seven benchmarks demonstrate that MultiSearch outperforms baseline methods, enhancing the SNR of retrieval and improving reasoning performance in question-answering tasks.
\end{abstract}
\section{Introduction}
\label{sec:introduction}

% % 1. Context - why deepsearch is important
% Deep search agents have shown substantial promise in addressing complex information retrieval and reasoning tasks\cite{deepsearch-agents, llm-base-deepsearch,webthinker,tongyi-deepsearch}. 
% Built upon large language models (LLMs) as the cognitive backbone, these agents integrate reasoning with tool use in a unified framework~\cite{llm-tool-learning}. 
% They iteratively generate reasoning traces and search actions, and adapt their strategies based on intermediate states, allowing dynamic acquisition of external knowledge from diverse large-scale sources to enhance their generation capabilities\cite{rag}.
% This synergy of reasoning and retrieval provides a practical pathway for tackling tasks that demand both structured reasoning and real-time information retrieval.

% % 2. Review - existing deepsearch methods
% Most existing methods follow a ReAct~\cite{react}-style "retrieval-during-reasoning" paradigm~\cite{search-o1, r1-searcher, search-r1, zerosearch, research, autorefine}. 
% Specifically, the model generates search queries conditioned on its current reasoning state, retrieve the top-$k$ relevant documents from external knowledge sources, and update the internal state accordingly. 
% This iterative process continues until a final answer is produced. 
% Subsequently, the model is typically optimized via reinforcement learning (RL), using reward signals derived from outputs in designated blocks, enabling gradual improvement in its reasoning and interaction with search tools.
Large language models (LLMs) have demonstrated strong capabilities in understanding and reasoning\cite{llm-eval-survey}. However, they remain constrained on knowledge-intensive tasks by their reliance on static internal knowledge\cite{llm-hallucination,llm-medical}. Retrieval-augmented generation (RAG) alleviates this limitation by enabling LLMs to access external knowledge during generation~\cite{rag,rag-method}. For complex questions, a single retrieval step is often insufficient, since the information required for answering may depend on intermediate reasoning states. This motivates deep search agents, which retrieve external knowledge during multi-step reasoning and use the retrieved information to support subsequent reasoning~\cite{deepsearch-agents,llm-base-deepsearch,webthinker,tongyi-deepsearch}.

Most existing deep search methods follow a ReAct-style retrieval-during-reasoning paradigm~\cite{react,search-o1,r1-searcher,search-r1,zerosearch,research}. At each reasoning step, the agent generates a search query conditioned on its current reasoning state, retrieves the top-$k$ relevant documents from an external knowledge source, and incorporates the retrieved information into the context for subsequent reasoning. This process repeats until the agent has gathered sufficient information to produce a final answer. To optimize this multi-step process, recent reinforcement learning (RL)-based methods model the search engine as part of the environment and train the agent on sampled reasoning-and-search trajectories~\cite{search-r1,autorefine}. 
The policy is typically updated with outcome-level rewards, such as final-answer correctness, which encourages better query generation, tool use, and final question-answering performance.

\begin{figure}[t]
    \centering
    \includegraphics[width=\linewidth]{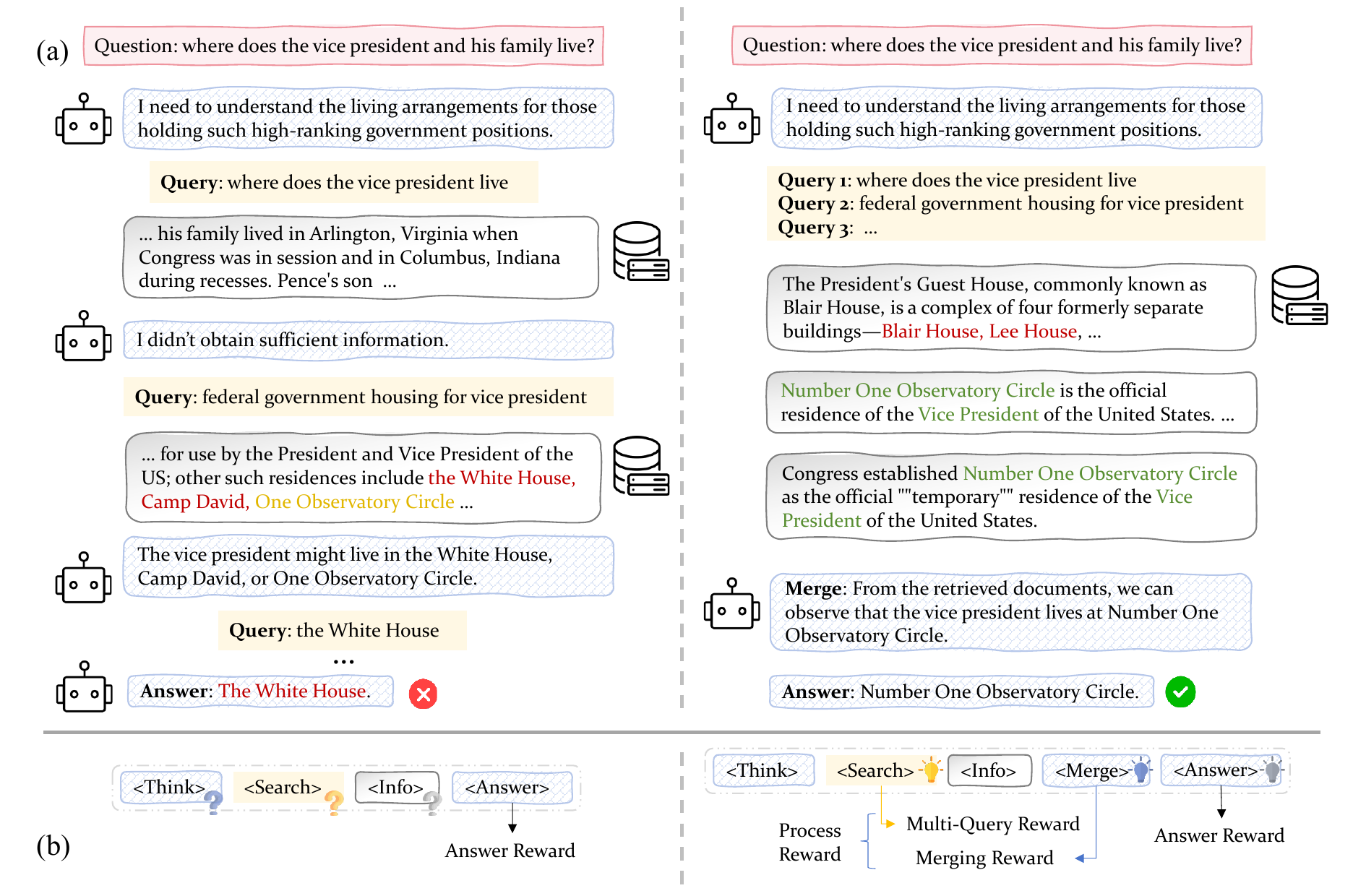}
    \caption{Comparison of classic deep search methods and MultiSearch. 
    (a) Previous serial single-query retrieval can be affected by noisy information and may consume more reasoning steps. MultiSearch employs multi-query retrieval with explicit merging to capture more comprehensive information with less steps.
    (b) Based on prior methods, MultiSearch incorporates more targeted rewards to enable fine-grained supervision.}
    \label{fig:teaser}
\end{figure}

% 3. Research Gaps - limitations of current deepsearch methods
Despite its effectiveness, this paradigm can limit the quality of the intermediate retrieval used for reasoning. In particular, we identify two key limitations:
\begin{itemize}[leftmargin=*]
    \item \textbf{Low Signal-to-Noise Ratio (SNR) of Retrieval.}
    At each reasoning step, existing methods typically rely on a single retrieval query, which captures only one formulation of the current information need. For multi-hop or complex questions involving multiple entities, relations, or sub-questions, this may retrieve only partial information. Moreover, if the generated query is under-specified, ambiguous, or mismatched with the corpus, the retrieved top-$k$ documents may include irrelevant or noisy content. The resulting lack of useful information and presence of noise can lower the signal-to-noise ratio (SNR) of the intermediate reasoning context, degrading reasoning accuracy or leading to unnecessary search steps, as illustrated by the single-query failures in Figure~\ref{fig:teaser}(a).
    \item \textbf{Underexplored fine-grained supervision.}
    % Early methods generally reward only the final answer, offering limited guidance for intermediate steps~\cite{search-r1}.
    % Subsequent works have introduced task-specific rewards, such as format rewards to encourage correct token generation and redundancy penalties to discourage repetitive retrievals\cite{r1-searcher}. 
    % Yet, there remain significant opportunities to explore more granular reward signals, particularly for dynamic reasoning and interaction, as shown in Figure~\ref{fig:teaser}(b). Such designs can better shape the reasoning trajectory and enhance system robustness.
    Recent RL-based deep search methods are often optimized with outcome-level rewards, such as final-answer correctness, sometimes supplemented by format or tool-use rewards~\cite{r1-searcher,search-r1,autorefine}. However, these rewards provide limited feedback on intermediate behaviors during retrieval-during-reasoning. In particular, when introducing intermediate mechanisms to improve retrieval quality and reduce noise, outcome-level supervision alone may be insufficient, as it provides limited guidance on whether the agent retrieves sufficient useful information and consolidates it into a reliable context for reasoning. This motivates a multi-process reward design that provides targeted supervision for both retrieval and information consolidation, as illustrated in Figure~\ref{fig:teaser}(b).
\end{itemize}

% 4. Propose: our method
% To address these challenges, we propose \textbf{MultiSearch}, a novel deep search framework with  parallel multi-route retrieval strategy and task-specific process supervision. First, the agent generates multiple search queries at each reasoning step, and issues them in parallel to the external knowledge base. The retrieved documents are then explicitly merged. By exploring multiple search routes concurrently, the agent accesses a broader set of information, reducing reliance on any single noisy retrieval path. The explicit merging step synthesizes key information from the aggregated results, improving the SNR of the retrieved content. Second, we extend the reward structure by introducing two auxiliary process reward signals: route count reward and merging reward, which are based on rule-based functions, eliminating the need for additional reward models. This design reduces computational overhead while enhancing interpretability. To effectively optimize the agent under this multi-reward setup, we adopt Group reward–Decoupled Normalization Policy Optimization (GDPO) algorithm~\cite{gdpo}, as it is well-suited for managing heterogeneous reward signals.
To address these limitations, we propose \textbf{MultiSearch}, an RL-based framework that improves retrieval-during-reasoning through multi-query retrieval and explicit merging of retrieved information. At each reasoning step, MultiSearch generates queries from multiple perspectives, including rephrasing, concept expansion, and question decomposition, and retrieves external information in parallel. This expands the scope of relevant information and reduces reliance on any single retrieval result. The agent then consolidates and refines the retrieved information through an explicit merging step, improving the SNR of the intermediate retrieval for subsequent reasoning.
To train the agent to use these mechanisms effectively, we introduce a multi-process reward design for reinforcement learning. In addition to the outcome reward for final-answer correctness, we use a multi-query reward to encourage retrieval with multiple queries and a merging reward to encourage information consolidation for high SNR. 
These process-level rewards provide targeted supervision for the intermediate retrieval and merging behaviors. 
We optimize the resulting multi-reward objective with Group reward-Decoupled Normalization Policy Optimization (GDPO)~\cite{gdpo}, which separately normalizes heterogeneous reward signals before aggregating them for policy optimization.

We train MultiSearch with Qwen2.5-3B/7B Base and Instruct backbones, and evaluate it on seven QA benchmarks, including three single-hop datasets~\cite{nq,popqa,triviaqa} and four multi-hop datasets~\cite{hotpotqa,2wikiqa,musique,bamboogle}. MultiSearch achieves the best average performance among the compared baselines~\cite{search-o1,search-r1,research,autorefine} across these settings. Experimental results show that the agent progressively learns to issue multiple retrieval queries and produce higher-SNR merged information, suggesting that the proposed reward design encourages the intended intermediate behaviors. Ablation studies further verify the contributions of multi-query retrieval, explicit merging, and corresponding rewards to the overall performance.
\section{Method}

\begin{figure}[t]
    \centering
    \includegraphics[width=\linewidth]{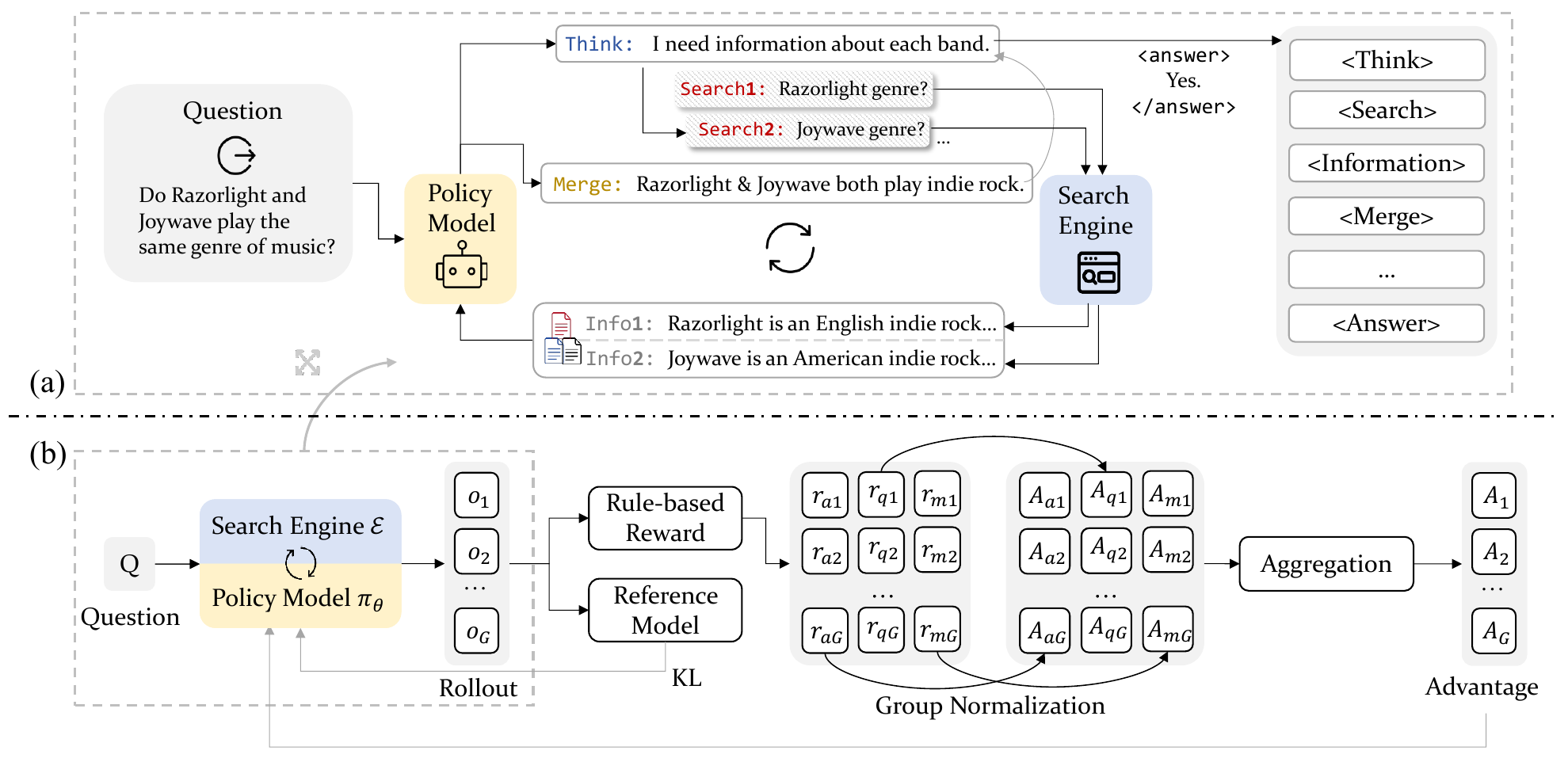}
    \caption{The training framwork of MultiSearch. (a) An question-answering example, including think, multi-query search, information, merge and answer steps. (b) Overview of the policy optimization procedure, where the model is trained using GDPO.}
    \label{fig:framework}
\end{figure}

In this section, we provide a comprehensive description of the MultiSearch framework. We begin by outlining the trajectory generation process, with a particular emphasis on our parallel multi-query retrieval mechanism (\S\ref{sec:multi-query}). 
Next, we present our multi-granularity reward design, which includes an answer reward, a multi-query reward, and a merging reward (\S\ref{sec:reward modeling}). 
Finally, we detail the training objective based on Group reward–Decoupled Normalization Policy Optimization(\S\ref{sec:rl}).

\subsection{Generation with Multi-Query Retrieval}
\label{sec:multi-query}
\paragraph{Rollout Generation.}
For each question $q$ in the training datasets, the agent iteratively interacts with the search engine $\mathcal{E}$ and generates a reasoning trajectory $o$ . Specifically, it generates multiple queries in \token{search}...\token{/search} to trigger retrieval tools, and uses \token{information}...\token{/information} to encapsulate the retrieved documents. Then, the agent explicitly extracts and merges key information from the retrieved content within \token{merge}...\token{/merge}. The existing response concatenated with the \token{information} and \token{merge} blocks serves as the next input prompt for the subsequent generation step. This \texttt{search$\to$info$\to$merge} cycle continues until the agent determines that there is sufficient information and presents an answer inside \token{answer}...\token{/answer}(\cf Figure \ref{fig:framework}(a)).

\paragraph{Multi-Query Retrieval.}
To scale the retrieval from diverse perspectives, we equip the agent with three query generation strategies: rephrasing, concept expansion, and question decomposition. Rephrasing helps retrieve documents that use different lexical or syntactic expressions, mitigating the risk of missing relevant information due to phrasing mismatches\cite{rephrase}. Concept expansion broadens the search scope by adding related terms, synonyms, or hypernyms, which is particularly useful when the initial query is too narrow\cite{concept-expan}. Question decomposition splits a complex question into simpler sub-questions, solves them in parallel, and then consolidates the retrieved information to produce the final answer. At each retrieval step, the agent generates three queries for parallel retrieval, adopting one or more of the above strategies or exploring alternative strategies autonomously.

\paragraph{Explicit Merging.}
After retrieval, repetitive or irrelevant documents are removed to eliminate redundancy. The agent reads the remaining documents and explicitly places relevant information within \token{merge} and \token{/merge}. The training template is illustrated in Appendix \ref{append:prompts}. For more analysis on different integration manners, please refer to Appendix~\ref{append:merge-modules}.

\subsection{Reward Modeling}
\label{sec:reward modeling}

The reward system of MultiSearch consists of three components: (1) Answer Reward, which evaluates the accuracy of the final prediction. (2) Multi-Query Reward, which encourages generating multiple queries for retrieval. (3) Merging Reward, which assesses the quality of the information consolidation.

\paragraph{Answer reward.}
We compute a word-level F1 score between the predicted answer enclosed in \token{answer}...\token{/answer} and the ground-truth answer to measure the correctness of the agent's prediction. The answer reward is defined as:
\begin{equation}
\label{eq:r_ans}
    r_{\text{ans}}=\text{F1}(a_{\text{pred}},a)=\frac{2n_{\text{int}}}{n_{\text{pred}}+n_{\text{truth}}}
\end{equation}
where $n_{\text{pred}}$ and $n_{\text{truth}}$ are the word counts of the predicted answer and the ground-truth answer, respectively. And $n_{\text{int}}$ is the word count of their intersection.

\paragraph{Multi-Query Reward.}
We introduce a dedicated reward for the proposed multi-query retrieval mechanism. Specifically, we extract all queries within the \token{search}...\token{/search} blocks throughout the rollout and calculate the average number of queries generated per step:
\begin{equation}
    r_{\text{query}}=\begin{cases}0.1, & n_\text{q}>2\\0, & \text{otherwise}\end{cases}
\end{equation}
where $n_{\text{q}}$ is the average number of queries per step.

\paragraph{Merging reward.}
The merging step is designed to remove irrelevant  information from the retrieved documents and to consolidate key evidence. To evaluate the quality of this integration, we aggregate all text enclosed within \token{merge}\token{/merge} blocks, and verify whether the ground-truth answer appears in any of them:
\begin{equation}
    r_{\text{merge}}=\begin{cases}0.1, & \exists \mathcal{M}_i\in \{\mathcal{M}_1,\mathcal{M}_2,...\mathcal{M}_n\},\mathcal{M}_i\cap a=a \\ 0, & \text{otherwise}\end{cases}
\end{equation}
where $\{\mathcal{M}_1,\mathcal{M}_2,...\mathcal{M}_n\}$ denotes the $n$ merging steps within a single rollout. The multi-query reward and merging reward are applied only when the answer is correct. In particular, the final rewards are defined as follows:
\begin{equation}
    (\mathcal{R}_{\text{ans}},\mathcal{R}_{\text{query}},\mathcal{R}_{\text{merge}})=\begin{cases}(r_{\text{ans}},r_{\text{query}},r_{\text{merge}}), & r_{\text{ans}}>0\\(0,0,0), & \text{otherwise}\end{cases}
\end{equation}

\subsection{Reinforcement Learning}
\label{sec:rl}
Group reward-Decoupled Normalization Policy Optimization (GDPO)\cite{gdpo} was proposed to address a key limitation of Group Relative Policy Optimization (GRPO)\cite{grpo} in multi‑reward RL. Unlike GRPO, which directly sums all reward components into a single rollout reward, GDPO normalizes each reward independently within the group. The resulting advantages are then aggregated and subjected to batch-wise normalization. This decoupled design preserves the distinct contributions of individual rewards, enabling the model to receive more fine‑grained and informative advantage signals.

We employ GDPO as the learning algorithm for RL training. Specifically, for each input question, given a policy model $\pi_\theta$ and a reference model $\pi_{\text{ref}}$, GDPO samples a group of rollouts $\{o_i\}_{i=1}^G$. The agent is optimized by maximizing the following objective:
\begin{equation}
    \begin{aligned}
    \mathcal{J}_{\text{GDPO}}(\theta) =  &\mathbb{E}_{x\sim \mathcal{D}, \{o_i\}_{i=1}^G \sim \pi_{\theta_{\text{old}}}(\cdot|x;\mathcal{E})}
    \Bigg[
    \frac{1}{G} \sum_{i=1}^{G} 
    \frac{1}{|o_i|} \sum_{t=1}^{|o_i|}
    \min \Bigg(
    \frac{ \pi_{\theta}(o_{i,t} \mid x, o_{i,<t};\mathcal{E}) }
     { \pi_{\theta_{\text{old}}}(o_{i,t} \mid x, o_{i,<t};\mathcal{E}) }
     \hat{A}_{i,t},
    \\ &
    \text{clip} \left(
    \frac{ \pi_{\theta}(o_{i,t} \mid x, o_{i,<t};\mathcal{E}) }
     { \pi_{\theta_{\text{old}}}(o_{i,t} \mid x, o_{i,<t};\mathcal{E}) }
    , 1-\epsilon, 1+\epsilon
    \right) \hat{A}_{i,t}
    \Bigg)
    - \beta \mathbb{D}_{\text{KL}} \left[ \pi_{\theta} \, \| \, \pi_{\text{ref}} \right]
    \Bigg]
    \end{aligned}
\end{equation}

where $G$ denotes the group size, $\mathcal{E}$ represents the search engine, $\epsilon$ is the clipping ratio and $\beta$ is the coefficient of KL divergence. $\hat{A}_{i,t}$ is the batch-wise normalized advantage for the $i$-th rollout in current group, defined as:
\begin{equation}
    \hat{A}_{i,j,t}=\frac{A_{i,j,t}-\mathrm{mean}\{A_{i',j',t}|i'\in D_\text{Batch},j'=1,...G\}}{\mathrm{std}\{A_{i',j',t}|i'\in D_\text{Batch},j'=1,...G\}}
\end{equation}
where
\begin{equation}
    A_{i,j,t}=\sum_{k\in\{\text{ans,query,merge}\}}w_k A^k_{i,j,t},\quad \text{with}\quad A^k_{i,j,t}= \frac{r^k_{i,j,t} - \mathrm{mean}(r^k_t)}{\mathrm{std}(r^k_t)}
\end{equation}
represents the weighted sum of the normalized advantages of different reward components. Figure \ref{fig:framework}(b) illustrates an overview of our GDPO-based training framework.
\section{Experiment}
In this section, we conduct a series of experiments to address the following research questions: (1) RQ1: How does MultiSearch perform on question-answering tasks? (2) RQ2: What are the contributions of the multi-query retrieval and explicit merging mechanism? (3) RQ3: Can our different query generation strategy effectively guide the search agent? (4) RQ4: How sensitive is MultiSearch to the number of retrieval queries and the retrieval depth? 

\subsection{Settings}
\paragraph{Datasets \& Evaluation Metrics.}
 We evaluate MultiSearch on three general Q\&A datasets: NQ\cite{nq}, TriviaQA\cite{triviaqa}, PopQA\cite{popqa}, and four multi-hop reasoning Q\&A datasets: HotpotQA\cite{hotpotqa}, 2WikiMultiHopQA (2Wiki)\cite{2wikiqa}, Musique\cite{musique}, and Bamboogle\cite{bamboogle}. For a fair comparison, we follow the settings of baseline methods\cite{search-r1,autorefine}, training the agent on a combined dataset of NQ and HotpotQA, and adopting Extract Match (EM) as the evaluation metric. For ablation on different evaluation metrics and training datasets, please refer to Appendix \ref{append:abla-metric} and \ref{append:abla-dataset}.
 
 \paragraph{Baselines.}
 To evaluate the effectiveness of our MultiSearch framework, we consider diverse baselines mainly grouped into three categories: (1) methods without retrieval: direct inference with LLM, Chain of Thought (CoT) reasoning\cite{cot}, Supervised Fine-Tuning (SFT), and R1-like fine-tuning\cite{deepseek-r1}. (2) methods with single-turn static retrieval: naive Retrieval-Augmented Generation (RAG)\cite{rag-method}. (3) methods with multi-turn dynamic retrieval: Search-o1\cite{search-o1}, Interleaving retrieval with chain-of-thought reasoning (IRCoT)\cite{ircot}, previous outstanding agentic RL methods including ReSearch\cite{research}, Search-R1\cite{search-r1} and AutoRefine\cite{autorefine}, as well as relatively recent work Dr.Zero\cite{drzero}, AdaSearch\cite{adasearch} and CriticSearch\cite{criticsearch}.
 
\begin{table}
    \caption{(\textbf{RQ1}) Main results. \textbf{Bold} denotes the best results, and \underline{underline} indicates the second best.}
    \label{tab:main-results}
    \centering
    \renewcommand{\arraystretch}{0.83} 
    \resizebox{\linewidth}{!}{ 
    \begin{tabular}{lllllllll}
      \toprule
      Methods & \multicolumn{3}{c}{Single-Hop QA} & \multicolumn{4}{c}{Multi-Hop QA} & \\
      \cmidrule{2-8}
         & NQ & TriviaQA & PopQA & HotpotQA & 2Wiki & Musique & Bamboogle & Avg. \\
      \midrule
      \multicolumn{9}{l}{\textbf{Qwen2.5-3B-Base/Instruct}} \\
      Direct Generation & 0.106 & 0.288 & 0.108 & 0.149 & 0.244 & 0.020 & 0.024 & 0.134 \\
      CoT & 0.023 & 0.032 & 0.005 & 0.021 & 0.021 & 0.002 & 0.000 & 0.015 \\
      IRCoT & 0.111 & 0.312 & 0.200 & 0.164 & 0.171 & 0.067 & 0.240 & 0.181 \\
      SFT & 0.249 & 0.292 & 0.104 & 0.186 & 0.248 & 0.044 & 0.112 & 0.176 \\
      R1-Instruct & 0.210 & 0.449 & 0.171 & 0.208 & 0.275 & 0.060 & 0.192 & 0.224 \\
      R1-Base & 0.226 & 0.455 & 0.173 & 0.201 & 0.268 & 0.055 & 0.224 & 0.229 \\
      RAG & 0.348 & 0.544 & 0.387 & 0.255 & 0.226 & 0.047 & 0.080 & 0.270 \\
      Search-o1 & 0.238 & 0.472 & 0.262 & 0.221 & 0.218 & 0.054 & 0.320 & 0.255 \\
      AdaSearch & 0.379 & 0.568 & 0.428 & 0.334 & 0.385 & 0.144 & 0.280 & 0.360 \\
      CriticSearch & - & - & - & 0.414 & 0.409 & 0.180 & 0.368 & - \\
      Dr.Zero & 0.391 & 0.572 & 0.431 & 0.298 & 0.291 & 0.091 & 0.200 & 0.326 \\ 	 	 	 	 	 	ReSearch-Instruct & 0.365 & 0.571 & 0.395 & 0.351 & 0.272 & 0.095 & 0.266 & 0.331 \\
      ReSearch-Base & 0.427 & 0.597 & 0.430 & 0.305 & 0.272 & 0.074 & 0.128 & 0.319 \\
      Search-R1-Instruct & 0.397 & 0.565 & 0.391 & 0.331 & 0.310 & 0.124 & 0.232 & 0.336 \\
      Search-R1-Base & 0.421 & 0.583 & 0.413 & 0.297 & 0.274 & 0.066 & 0.128 & 0.312 \\
      ZeroSearch-Instruct & 0.402 & 0.580 & \underline{0.460} & 0.228 & 0.214 & 0.104 & 0.181 & 0.310 \\
      ZeroSearch-Base & 0.434 & \textbf{0.638} & \textbf{0.484} & 0.322 & 0.356 & 0.138 & 0.153 & 0.361 \\
      StepSearch-Instruct & - & - & - & 0.345 & 0.320 & 0.174 & 0.344 & - \\
      StepSearch-Base & - & - & - & 0.329 & 0.339 & \underline{0.181} & 0.328 & - \\
      AutoRefine-Instruct & 0.436 & 0.597 & 0.447 & 0.404 & 0.380 & 0.169 & 0.336 & 0.396 \\
      AutoRefine-Base & 0.467 & 0.620 & 0.450 & 0.405 & 0.393 & 0.157 & 0.344 & 0.405 \\
      \rowcolor[gray]{0.9}
      MultiSearch-Instruct & \underline{0.470} & 0.615 & 0.444 & \underline{0.420} & \underline{0.412} & \textbf{0.183} & \underline{0.371} & \underline{0.416} \\
      \rowcolor[gray]{0.9}
      MultiSearch-Base & \textbf{0.471} & \underline{0.630} & 0.455 & \textbf{0.431} & \textbf{0.413} & 0.163 & \textbf{0.390} & \textbf{0.422} \\
      \midrule
      \multicolumn{9}{l}{\textbf{Qwen2.5-7B-Base/Instruct}} \\
      Direct Inference & 0.134 & 0.408 & 0.140 & 0.183 & 0.250 & 0.031 & 0.120 & 0.181 \\
      CoT & 0.048 & 0.185 & 0.054 & 0.092 & 0.111 & 0.022 & 0.232 & 0.106 \\
      IRCoT & 0.224 & 0.478 & 0.301 & 0.133 & 0.149 & 0.072 & 0.224 & 0.239 \\
      SFT & 0.318 & 0.354 & 0.121 & 0.217 & 0.259 & 0.066 & 0.112 & 0.207 \\
      R1-instruct & 0.270 & 0.537 & 0.199 & 0.237 & 0.292 & 0.072 & 0.293 & 0.271 \\
      R1-base & 0.297 & 0.539 & 0.202 & 0.242 & 0.273 & 0.083 & 0.296 & 0.276 \\
      RAG & 0.349 & 0.585 & 0.392 & 0.299 & 0.235 & 0.058 & 0.208 & 0.304 \\
      Search-o1 & 0.151 & 0.443 & 0.131 & 0.187 & 0.176 & 0.058 & 0.296 & 0.206 \\ 
      ReasonRAG & - & - & 0.415 & 0.384 & \textbf{0.436} & 0.128 & 0.360 & - \\
      CtriticSearch & - & - & - & 0.442 & \underline{0.428} & 0.194 & \underline{0.472} & - \\
      Dr.Zero & 0.406 & 0.608 & 0.416 & 0.362 & 0.347 & 0.104 & 0.360 & 0.372 \\
      Search-R1-Instruct & 0.429 & 0.623 & 0.427 & 0.386 & 0.346 & 0.162 & 0.400 & 0.396 \\
      Search-R1-Base & 0.395 & 0.560 & 0.388 & 0.326 & 0.297 & 0.125 & 0.360 & 0.350 \\
      ZeroSearch-Instruct & 0.414 & 0.574 & 0.448 & 0.274 & 0.300 & 0.098 & 0.111 & 0.317 \\	 
      ZeroSearch-Base & 0.430 & 0.616 & 0.414 & 0.338 & 0.346 & 0.130 & 0.139 & 0.345 \\	 
      StepSearch-Instruct & - & - & - & 0.386 & 0.366 & \textbf{0.226} & 0.312 & - \\		 
      StepSearch-Base & - & - & - & 0.380 & 0.385 & \underline{0.216} & 0.467 & - \\
      AutoRefine-Instruct & 0.419 & 0.618 & 0.403 & 0.408 & 0.325 & 0.188 & 0.416 & 0.397 \\
      AutoRefine-Base & 0.470 & \underline{0.649} & \textbf{0.463} & \underline{0.445} & 0.369 & 0.204 & 0.435 & \underline{0.434} \\
      \rowcolor[gray]{0.9}
      MultiSearch-Instruct & \underline{0.472} & 0.636 & 0.437 & 0.421 & 0.351 & 0.171 & 0.463 & 0.422 \\
      \rowcolor[gray]{0.9}
      MultiSearch-Base & \textbf{0.491} & \textbf{0.657} & \underline{0.458} & \textbf{0.446} & 0.416 & 0.170 & \textbf{0.476} & \textbf{0.445} \\
      \bottomrule
    \end{tabular}
    }
\end{table}

\paragraph{Implementation details.}
In line with previous work, we use Qwen2.5-3B-Base/Instruct and Qwen2.5-7B-Base/Instruct as the backbone models. For retrieval, we utilize E5\cite{e5} as the search engine and 2018 Wikipedia dump\cite{wiki-2018} as the external data source. Both the number of queries and retrieved documents are set to 3. 
% The reported results are the average of three tests, and the error is small so we omit the report of the variance. 
The reported results are averaged over three runs, and we omit the variance as it is relatively small.
Most of the baseline results are taken directly from the corresponding original papers or paper of Search-R1 under comparable settings, while AutoRefine is reproduced using open-source code. Additional details of the experimental settings can be found in Appendix \ref{append:train-detail}.

\subsection{Main Results (RQ1)}

\textbf{Overall Performance.}
The performance of MultiSearch compared to baseline methods is presented in Table~\ref{tab:main-results}. As shown, MultiSearch achieves the highest average accuracy across seven benchmarks (column \textit{Avg.}) for both 3B and 7B model sizes. The gains are particularly notable on multi-hop benchmarks, where multi-query retrieval expands the scope of relevant information to cover multiple entities, relations, or sub-questions, and explicit merging consolidates the retrieved information for subsequent reasoning. Additionally, base model variants outperform their instruction-tuned counterparts. As discussed in Appendix~\ref{append:it-base}, one possible explanation is that instruction fine-tuning may reduce the model's adaptability to new multi-step reasoning tasks. 

\begin{figure}
    \centering
    \begin{minipage}{0.58\linewidth}
        \centering
        % \vspace{-4mm}
          \subfloat[]
          {
              \label{fig:snr}\includegraphics[width=0.49\textwidth]{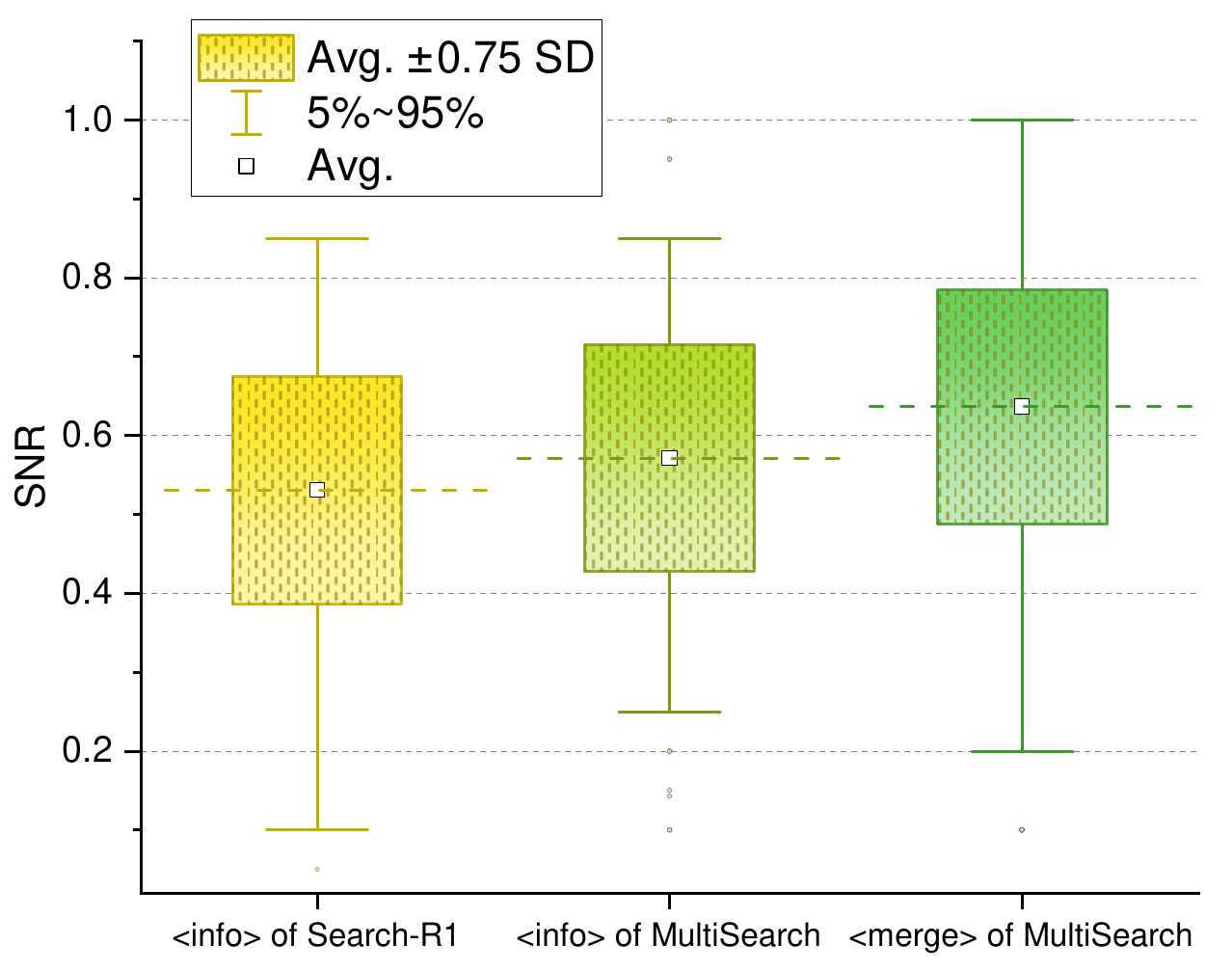}
          }
          \subfloat[]
          {
              
              \label{fig:avg-search-col}\includegraphics[width=0.49\textwidth]{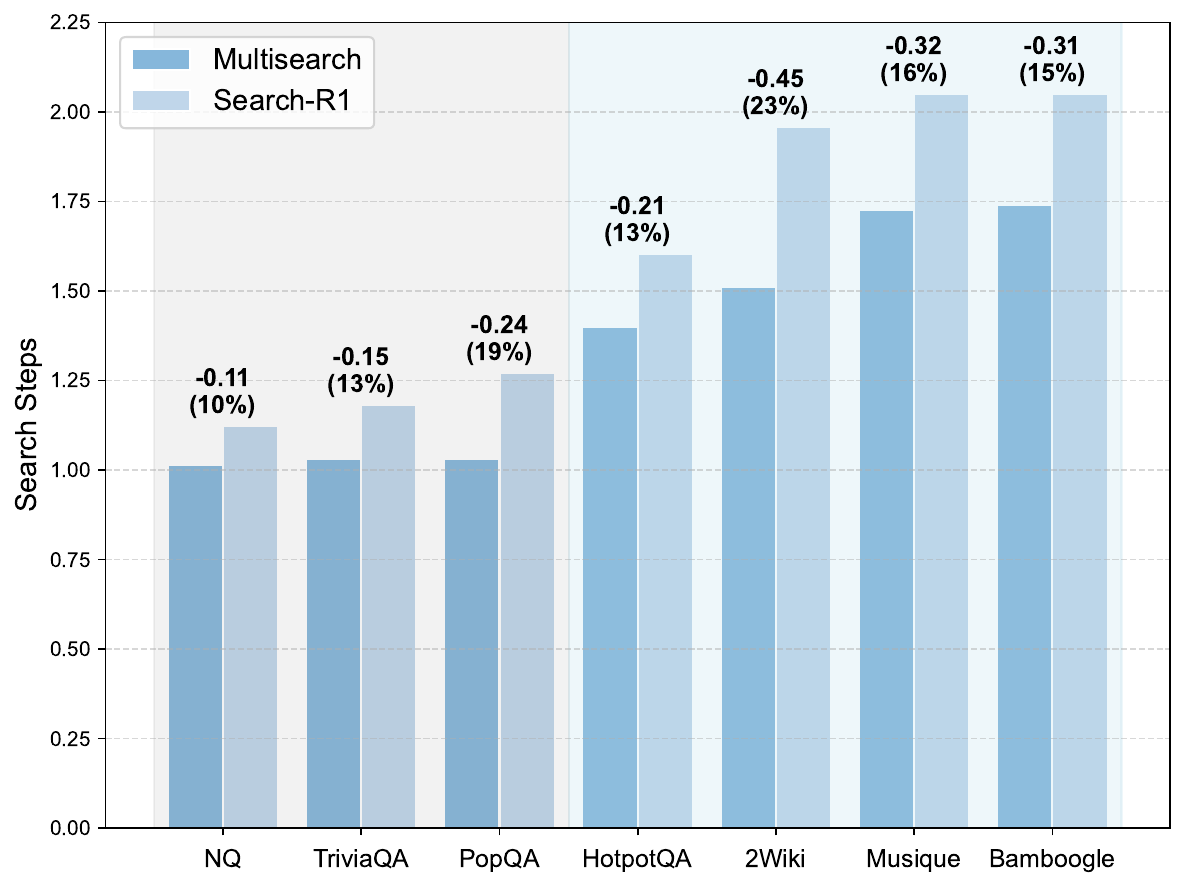}
          }
          \caption{Comparison of the retrieval quality and efficiency between Search-R1 and MultiSearch. (a) Signal to noise ratio of \token{information} and \token{merge} blocks. (b) Average number of search steps across different datasets. }
          \label{fig:turns-snr}
    \end{minipage}\hspace{8pt}
    \begin{minipage}{0.33\linewidth}
        \centering
        \includegraphics[width=\linewidth]{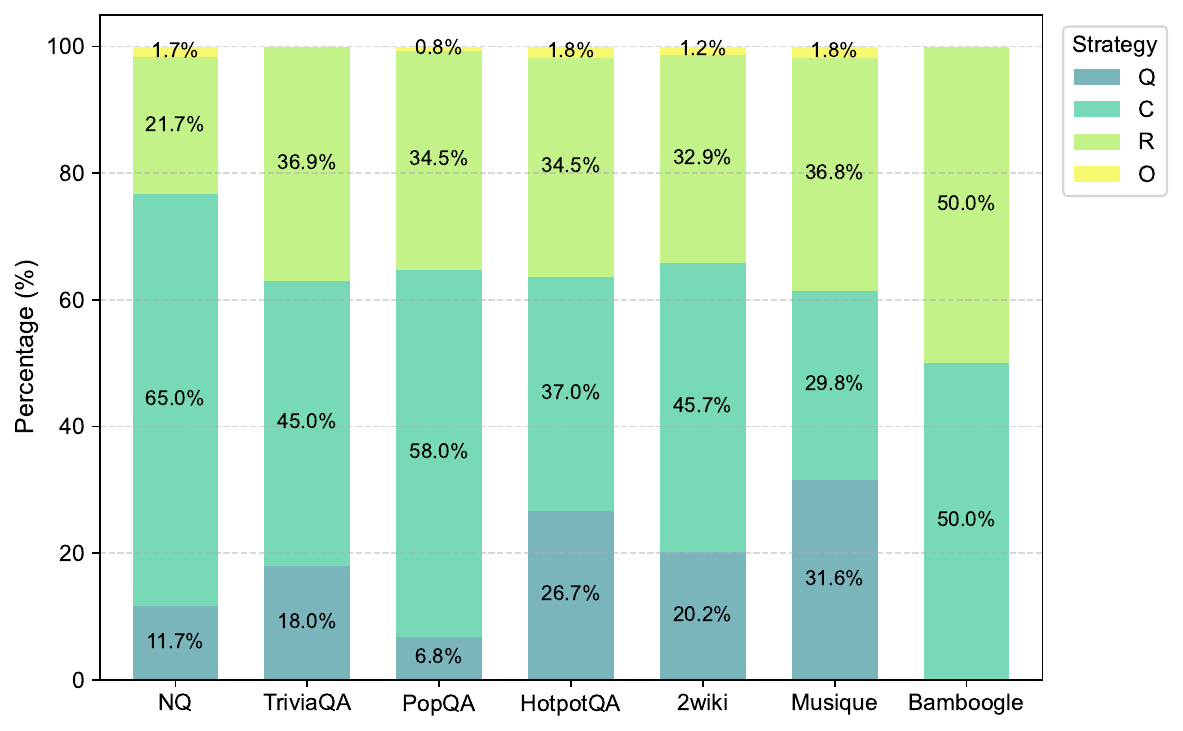}
        \caption{Distribution of various query generation strategies: question decomposition (Q), concept expansion (C), rephrasing (R), and others (O).}
        \label{fig:query-cate}
        \vspace{-0.45cm}
    \end{minipage}
    % \vspace{-0.5cm}
\end{figure}

\textbf{Retrieval Quality and Reasoning Steps.}
Figure~\ref{fig:turns-snr}(a) first examines the SNR of intermediate retrieved information. MultiSearch produces higher-SNR \texttt{<information>} blocks than Search-R1, and the explicit \texttt{<merge>} step further improves the SNR. This indicates that multi-query retrieval broadens useful information coverage, while merging helps reduce noisy content and forms a more reliable context for subsequent reasoning. This higher-SNR information helps reduce the need for follow-up searches to compensate for incomplete or noisy retrieval, consistent with the fewer search steps observed in Figure~\ref{fig:turns-snr}(b).

\subsection{Ablation on Key Components (RQ2)}
To assess the contributions of key components in MultiSearch, we perform ablation studies on Qwen2.5-3B-Instruct, with results reported in Table \ref{abla:key-components}. Specifically, we define five variants: (1) the full MultiSearch, (2) MultiSearch without merging reward (w/o $\mathcal{R}_{\text{merge}}$), (3) MultiSearch without both merging reward and multi-query reward (w/o $\mathcal{R}_{\text{merge}}$ \& $\mathcal{R}_{\text{query}}$), (4) MultiSearch further excluding explicit merging (w/o $\mathcal{R}_{\text{merge}}$ \& $\mathcal{R}_{\text{query}}$ \& \token{merge}), and (5) MultiSearch reduced to single-query retrieval with only answer reward (w/o all). The results demonstrate that both the multi-query retrieval \& merging step, as well as our reward modeling, contribute positively to the overall performance.
As shown in Figure \ref{fig:gr-vs-gd}, the merging reward $\mathcal{R_{\text{merge}}}$ and multi-query reward $\mathcal{R}_{\text{query}}$ gradually converge with the answer reward $\mathcal{R}_\text{ans}$ during training. This observation suggests that the model progressively learns to generate multiple queries and consolidate key evidence from the retrieved documents to enhance its question-answering performance.

\begin{table}
    \caption{(\textbf{RQ2}) Ablation on multi-query retrieval, merging operation, and corresponding rewards.}
    \label{abla:key-components}
    \centering
    \resizebox{\columnwidth}{!}{ 
    \begin{tabular}{lllllllll}
      \toprule
      Variants & \multicolumn{3}{c}{Single-Hop QA} & \multicolumn{4}{c}{Multi-Hop QA} & \\
      \cmidrule{2-8}
         & NQ & TriviaQA & PopQA & HotpotQA & 2Wiki & Musique & Bamboogle & Avg. \\
      \midrule
      \rowcolor[gray]{0.9}
      MultiSearch & \textbf{0.470} & \textbf{0.615} & \textbf{0.444} & \textbf{0.420} & \textbf{0.412} & \textbf{0.183} & \underline{0.371} & \textbf{0.416} \\
      w/o $\mathcal{R}_{\text{merge}}$ & \underline{0.449} & \textbf{0.615} & \underline{0.434} & \underline{0.395} & \underline{0.405} & 0.152 & \textbf{0.379} & \underline{0.404} \\
      w/o $\mathcal{R}_{\text{merge}}$ \& $\mathcal{R}_{\text{query}}$ & 0.442 & 0.606 & 0.420 & 0.396 & 0.394 & 0.152 & 0.368 & 0.397 \\
      w/o $\mathcal{R}_{\text{merge}}$ \& $\mathcal{R}_{\text{query}}$ \& \token{merge} & 0.443 & 0.597 & 0.425 & 0.393 & 0.328 & \underline{0.163} & 0.350 & 0.386 \\
      w/o all & 0.427 & 0.576 & 0.424 & 0.378 & 0.344 & 0.157 & 0.315 & 0.374 \\
      \bottomrule
    \end{tabular}
    }
\end{table}

% \begin{figure}
%     \centering
%     \includegraphics[width=\linewidth]{fig/reward-valid-length-3b.pdf}
%     \caption{Analysis of training dynamics.}
%     \label{fig:reward-valid-length}
% \end{figure}

\subsection{Ablation on query generation strategies (RQ3)}
To address RQ3, we compare the performance under different query generation strategies. In Table \ref{abla:query-enhance}, ``w/ Rephrase'', ``w/ Concept'', and ``w/ Decompose'' represent scenarios where the agent generate multiple queries merely under a single strategy: rephrasing, concept expansion, or question decomposition, respectively. ``Simple'' refers to a setting where only multi-query generation is required, without any strategies as guidance. As illustrated in Table \ref{abla:query-enhance}, the agent guided by various strategies outperforms the one guided by a single strategy. This suggests that the multi-perspective strategies provide complementary benefits: rephrasing helps reduce lexical mismatch, concept expansion broadens the search scope, and question decomposition helps handle complex information needs.  Figure~\ref{fig:query-cate} further shows that the trained agent uses a mixture of strategies across datasets, supporting the need for multi-perspective query generation in retrieval-during-reasoning.

\begin{table}
    \caption{(\textbf{RQ3}) Ablation study on different query generation strategies.}
    \label{abla:query-enhance}
    \centering
    \renewcommand{\arraystretch}{0.8} 
    \resizebox{\columnwidth}{!}{ 
    \begin{tabular}{lllllllll}
      \toprule
      Variants & \multicolumn{3}{c}{Single-Hop QA} & \multicolumn{4}{c}{Multi-Hop QA} & \\
      \cmidrule{2-8}
         & NQ & TriviaQA & PopQA & HotpotQA & 2Wiki & Musique & Bamboogle & Avg. \\
      \midrule
      \rowcolor[gray]{0.9}
      MultiSearch & \textbf{0.470} & \textbf{0.615} & \textbf{0.444} & \textbf{0.420} & \textbf{0.412} & \textbf{0.183} & \underline{0.371} & \textbf{0.416} \\
      w/ rephrase & 0.437 & 0.606 & \underline{0.423} & 0.395 & 0.390 & 0.165 & 0.363 & \underline{0.397} \\
      w/ concept & 0.436 & 0.609 & 0.415 & \underline{0.398} & 0.355 & \underline{0.168} & \textbf{0.379} & 0.394 \\
      w/ decompose & \underline{0.442} & \underline{0.612} & 0.412 & 0.383 & \underline{0.396} & 0.142 & 0.320 & 0.387 \\
      simple & 0.418 & 0.604 & 0.421 & 0.373 & 0.375 & 0.136 & 0.320 & 0.378 \\
      \bottomrule
    \end{tabular}
    }
\end{table}

\subsection{Sensitivity Analysis (RQ4)}
\label{sec:para-sensi}
\begin{figure}
    \centering
    \includegraphics[width=\linewidth]{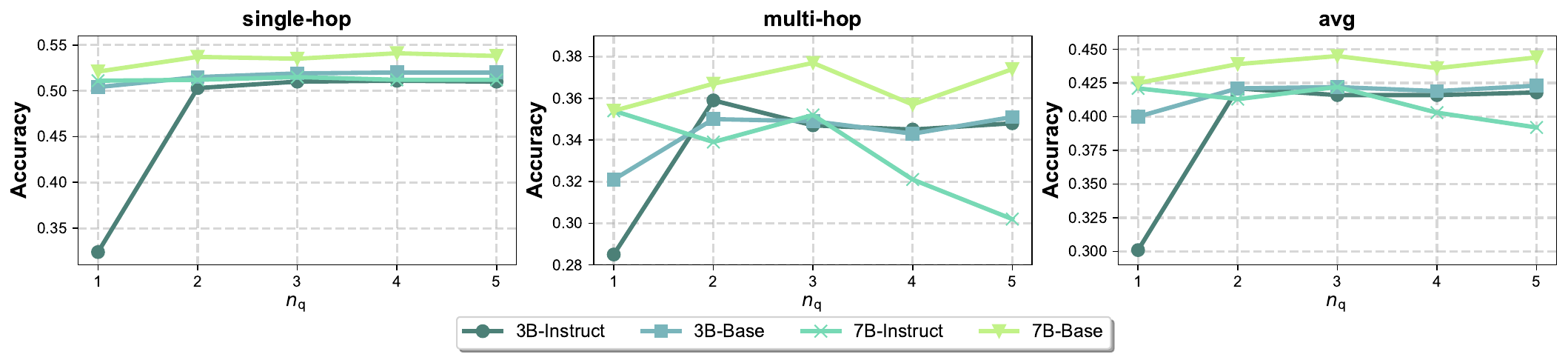}
    \caption{(\textbf{RQ4}) Performance of MultiSearch on different $n_\text{q}$, demonstrating the sensitivity of MultiSearch to the
number of retrieval queries.}
    \label{fig:para-n_query}
    \includegraphics[width=\linewidth]{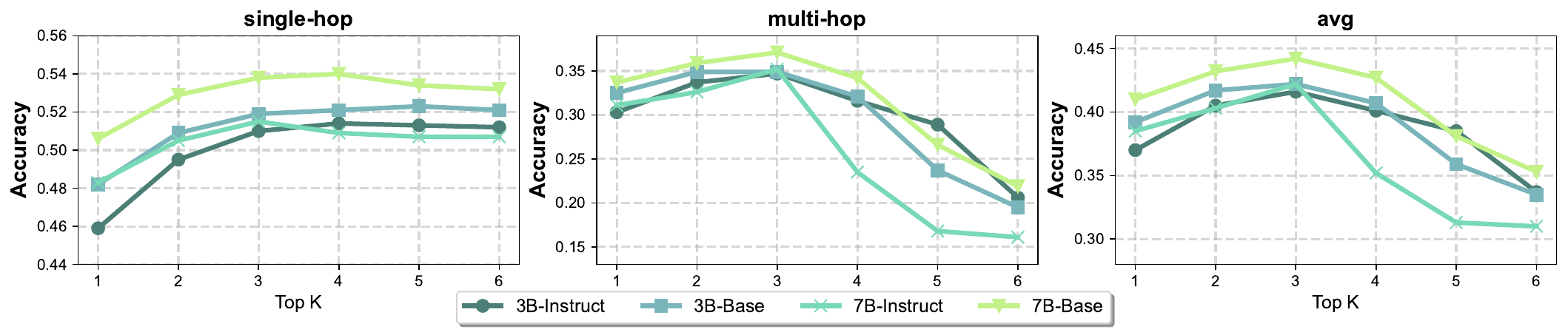}
    \caption{(\textbf{RQ4}) Performance of MultiSearch on different $\text{Top k}$, demonstrating the sensitivity of MultiSearch to the
retrieval depth.}
    \label{fig:para-topk}
    \vspace{-4mm}
\end{figure}

Different hyperparameters may affect the performance of deep search agents. We investigate the sensitivity of MultiSearch to the number of queries $n_\text{q}$ and the retrieval depth $k$. The results are presented in Figure \ref{fig:para-n_query} and \ref{fig:para-topk}, where ``single-hop'', ``multi-hop'', ``avg'' denote the average accuracy on single-hop benchmarks, multi-hop benchmarks, and all seven benchmarks, respectively. As shown in Figure \ref{fig:para-n_query}, performance peaks when $n_\text{q}=3$. Using too few queries fails to cover sufficient information, whereas too many queries may lead to supersaturation, introducing repetitive or irrelevant information. As illustrated in Figure \ref{fig:para-topk}, performance improves as $k$ increases when $k\leq 3$, but begins to decline when $k\geq 4$, as the search engine starts returning less relevant documents.

\subsection{Different RL Methods}
\begin{figure}
    \centering
    \includegraphics[width=\linewidth]{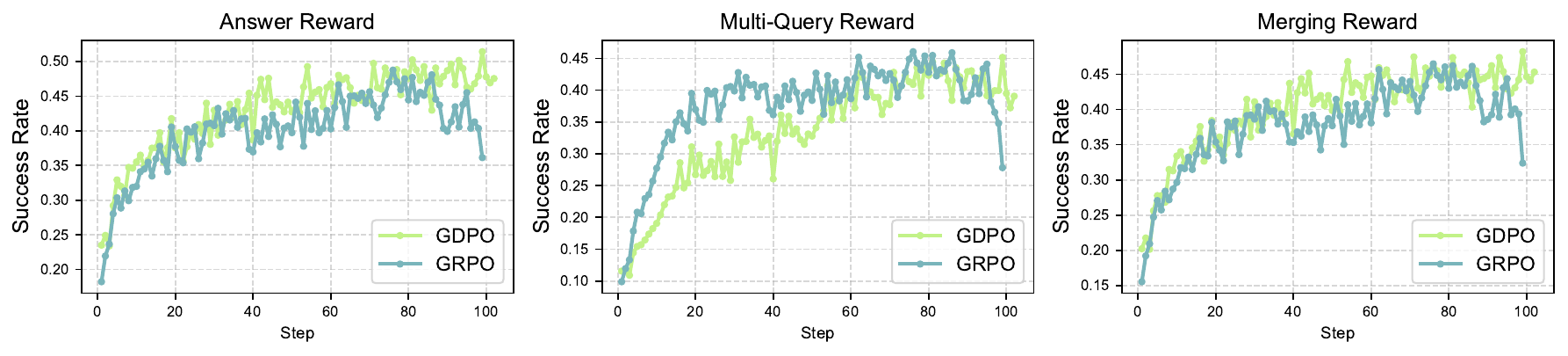}
     % \vspace{-4mm}
    \caption{Analysis of training rewards under GRPO and GDPO.}
    \label{fig:gr-vs-gd}
    \vspace{-4mm}
\end{figure}

We further evaluate MultiSearch using GRPO and GDPO as RL algorithms. The evaluation results are presented in Table \ref{tab:gr-vs-gd}, and the training dynamics is shown in Figure \ref{fig:gr-vs-gd}. As illustrated in the Figure, both methods successfully optimize the three different objectives, guiding the agent to follow the multi-query retrieval mechanism and merge key evidence during training. However, compared with GDPO, GRPO exhibits less balanced reward optimization. The multi-query reward rises quickly in the early stage (\cf Figure~\ref{fig:gr-vs-gd}(b)), while the answer and merging rewards improve more slowly (\cf Figure~\ref{fig:gr-vs-gd}(a)(c)). Since the multi-query reward is easier to satisfy than rewards related to answer correctness or information consolidation, it may dominate the aggregated reward signal under GRPO. GDPO mitigates this issue by normalizing reward components separately, resulting in more robust optimization and better performance, as shown in Table~\ref{tab:gr-vs-gd}.

\begin{table}
    \caption{Comparison between GRPO and GDPO on different backbone models.}
    \label{tab:gr-vs-gd}
    \centering
    \resizebox{\columnwidth}{!}{ 
    \begin{tabular}{lllllllll}
      \toprule
      \multirow{2}{*}{Methods} & \multicolumn{3}{c}{Single-Hop QA} & \multicolumn{4}{c}{Multi-Hop QA} & \\
      \cmidrule{2-8}
         & NQ & TriviaQA & PopQA & HotpotQA & 2Wiki & Musique & Bamboogle & Avg. \\
      \midrule
      \multicolumn{9}{l}{\textbf{Qwen2.5-3B-Base/Instruct}} \\
      MultiSearch-Base (GDPO) & 0.471 & 0.630 & 0.455 & 0.431 & 0.413 & 0.163 & 0.390 & 0.422 \\
      MultiSearch-Instruct (GDPO) & 0.470 & 0.615 & 0.444 & 0.420 & 0.412 & 0.183 & 0.371 & 0.416 \\
       \hdashline
      MultiSearch-Base (GRPO) & 0.450 & 0.606 & 0.435 & 0.399 & 0.393 & 0.165 & 0.372 & 0.403 \\
      MultiSearch-Instruct (GRPO) & 0.426 & 0.605 & 0.424 & 0.380 & 0.366 & 0.140 & 0.344 & 0.384 \\
       \midrule
       \multicolumn{9}{l}{\textbf{Qwen2.5-7B-Base/Instruct}} \\
      MultiSearch-Base (GDPO) & 0.491 & 0.657 & 0.458 & 0.446 & 0.416 & 0.170 & 0.476 & 0.445 \\
      MultiSearch-Instruct (GDPO) & 0.472 & 0.636 & 0.437 & 0.421 & 0.351 & 0.171 & 0.463 & 0.422 \\
       \hdashline
      MultiSearch-Base (GRPO) & 0.458 & 0.617 & 0.432 & 0.389 & 0.343 & 0.163 & 0.419 & 0.403 \\
      MultiSearch-Instruct (GRPO) & 0.418 & 0.637 & 0.424 & 0.403 & 0.350 & 0.158 & 0.415 & 0.400 \\
      \bottomrule
    \end{tabular}
    }
\end{table}
\section{Related Work}
\label{sec:related-work}
\paragraph{Deep Search Agents.}  
LLMs have demonstrated strong reasoning capabilities~\cite{cot,deepseek-r1,gpt4,qwen2_5-math,qwen2_5-coder}, but still suffer from  hallucinations and limited knowledge~\cite{llm-hallucination}. Retrieval-augmented generation (RAG) alleviates these issues by incorporating external knowledge~\cite{rag,rag-method}. However, native single-turn RAG models struggle with complex questions, as they lack mechanisms to assess evidence sufficiency or perform iterative retrieval. Prompt-based multi-turn methods partially address this limitation, yet remain suboptimal in optimizing tool-call decisions~\cite{react,ircot}. Supervised fine-tuning (SFT) effectively improves tool usage, but typically require large-scale labeled data, which is costly to construct and maintain~\cite{toolformer,self-rag}.
Recent work has explored reinforcement learning for deep search agents. Search-R1 models the search engine as part of the environment and uses outcome-level rewards to guide search behavior~\cite{search-r1}. R1-Searcher introduces a two-stage training framework that separately improves search-call formatting and answer accuracy~\cite{r1-searcher}. Subsequent methods, such as AutoRefine and EviNote-RAG, further enhance retrieval-augmented reasoning with post-retrieval processing or task-specific rewards~\cite{autorefine,evinote}. Other methods investigate principle-based reward models to provide supervision for intermediate steps~\cite{ppr,reasonrag}.
However, most previous methods rely on single-query retrieval, limiting the utility of each retrieval step \cite{zerosearch,drzero,searchr2}. In contrast, our MultiSearch leverages parallel multi-query retrieval and explicit merging to improve the quality of the search process.

\paragraph{Reinforcement Learning.}  
Reinforcement Learning from Human Feedback (RLHF) represents an early approach that transitions models from purely imitative behavior to strategic exploration~\cite{rlhf,rl-survey}. Proximal Policy Optimization (PPO)~\cite{ppo} is widely used due to its stable learning dynamics, but it can incur high computational costs. To mitigate this, several simplified variants such as Direct Preference Optimization (DPO)~\cite{dpo} and Group Relative Policy Optimization (GRPO)~\cite{grpo} have been proposed. 
While early RL methods primarily target preference alignment, recent works explore RL in retrieval-augmented reasoning settings. Both PPO and GRPO have demonstrated effectiveness in these scenarios. Considering both computational efficiency and performance, we adopt Group reward-Decoupled Normalization Policy Optimization (GDPO)~\cite{gdpo}, an improved variant of GRPO specifically designed to handle multi-reward objectives.
\section{Conclusion}
\label{conclusion}
% In this work, we introduce MultiSearch, a novel agentic reinforcement learning framework designed to enhance the search-during-think capability of deep search agents. MultiSearch incorporates a multi-route retrieval strategy to facilitate the acquisition of high-signal, low-noise information. By combining outcome-based rewards with auxiliary step-specific rewards for policy optimization, the agent is encouraged to gather more comprehensive information and extract relevant parts for reasoning. Experimental results show that MultiSearch improves answer accuracy, achieving up to a $x\%$ average improvement on standard benchmark datasets.
% Future work will aim to address the limitations discussed above. First, extending the evaluation to a broader range of tasks beyond QA datasets would help better assess the versatility of MultiSearch across more diverse domains. Second, integrating online search engines could enable more dynamic and real-time information access, which may improve the system's adaptability to continuously evolving data sources. Finally, exploring more flexible and adaptive query generation strategies remains a promising direction for improving retrieval quality. 
% Overall, these directions may further improve the generality of the proposed framework and facilitate its application in more realistic settings.
In this work, we introduce \textbf{MultiSearch}, an RL-based framework for improving retrieval-during-reasoning in deep search agents. Instead of relying on a single query at each reasoning step, MultiSearch generates queries from multiple perspectives and retrieves external information in parallel, expanding the scope of relevant information and reducing reliance on any single retrieval result. The retrieved information is then explicitly merged, allowing the agent to consolidate useful information and reduce noise before subsequent reasoning. To optimize these intermediate behaviors, we further propose a multi-process reward design that supervises both multi-query retrieval and information consolidation. Experiments on seven question-answering benchmarks show that MultiSearch improves the SNR of retrieved information and achieves better reasoning performance than strong baseline methods. These results suggest that improving both retrieval coverage and information consolidation is important for building more effective deep search agents.

\newpage
\bibliographystyle{unsrtnat}
\bibliography{custom}

\newpage
\appendix

\section{More Experimental Results}
\subsection{Ablation of Evaluation Metrics}
\label{append:abla-metric}

To provide a more comprehensive evaluation, we further assess MultiSearch-3B-Base using additional metrics, including F1 score (F1) and Coverage Exact Match (CEM). As shown in Table~\ref{fig:abla-metrics}, MultiSearch consistently outperforms the baseline methods Search-R1 and AutoRefine across all evaluated metrics, verifying the effectiveness of our framework.

\begin{table}[ht]
    \caption{Ablation of Evaluation Metrics.}
    \label{fig:abla-metrics}
    \centering
    \resizebox{\linewidth}{!}{ 
    \begin{tabular}{llllllllll}
      \toprule
      \multirow{2}{*}{Metrics} & \multirow{2}{*}{Methods} & \multicolumn{3}{c}{Single-Hop QA} & \multicolumn{4}{c}{Multi-Hop QA} & \\
      \cmidrule{3-9}
        & & NQ & TriviaQA & PopQA & HotpotQA & 2Wiki & Musique & Bamboogle & Avg. \\
      \midrule
      \multirow{3}{*}{EM} & Search-R1 & 0.421 & 0.583 & 0.413 & 0.297 & 0.274 & 0.066 & 0.128 & 0.312 \\
           & AutoRefine & 0.467 & 0.620 & 0.450 & 0.405 & 0.393 & 0.157 & 0.344 & 0.405 \\
           & MultiSearch & \textbf{0.471} & \textbf{0.630} & \textbf{0.455} & \textbf{0.431} & \textbf{0.413} & \textbf{0.163} & \textbf{0.390} & \textbf{0.422} \\
      \midrule
      \multirow{3}{*}{F1} & Search-R1 & 0.476 & 0.650 & 0.429 & 0.380 & 0.322 & 0.123 & 0.184 & 0.366 \\
           & AutoRefine & 0.534 & 0.689 & 0.479 & 0.503 & 0.453 & 0.233 & 0.449 & 0.477 \\
           & MultiSearch & \textbf{0.564} & \textbf{0.703} & \textbf{0.497} & \textbf{0.544} & \textbf{0.483} & \textbf{0.256} & \textbf{0.542} & \textbf{0.513} \\
      \midrule
      \multirow{3}{*}{CEM} & Search-R1 & 0.462 & 0.642 & 0.442 & 0.325 & 0.288 & 0.082 & 0.128 & 0.338 \\
           & AutoRefine & 0.502 & 0.674 & 0.468 & 0.440 & 0.428 & 0.175 & 0.384 & 0.439 \\
           & MultiSearch & \textbf{0.514} & \textbf{0.679} & \textbf{0.471} & \textbf{0.466} & \textbf{0.458} & \textbf{0.181} & \textbf{0.433} & \textbf{0.457} \\
      \bottomrule
    \end{tabular}
    }
\end{table} 

\subsection{Ablation of Training Datasets}
\label{append:abla-dataset}

MultiSearch is trained on a combined dataset of single-hop NQ and multi-hop HotpotQA, following prior work settings. To examine the effectiveness of this design, we further conduct experiments by training on single-hop data only (NQ) and multi-hop data only (HotpotQA), respectively. The results are reported in Table~\ref{tab:abla-train-data}. \textit{S-avg} and \textit{M-avg} denotes the average accuracy over three single-hop datasets and four multi-hop datasets.
We observe that the joint training setting achieves the best overall performance. In contrast, training solely on NQ leads to degraded performance on multi-hop tasks, while training solely on HotpotQA results in lower performance on single-hop tasks. These suggest that joint training on both types of data provides a more balanced learning signal across different reasoning settings.

\begin{table}[htbp]
    \caption{Comparison between different types of training datasets.}
    \label{tab:abla-train-data}
    \centering
    \resizebox{\columnwidth}{!}{ 
    \begin{tabular}{lllllllllll}
      \toprule
      \multirow{2}{*}{Training Datasets} & \multicolumn{4}{c}{Single-Hop QA} & \multicolumn{5}{c}{Multi-Hop QA} & \\
      \cmidrule{2-10}
         & NQ & TriviaQA & PopQA & S-Avg. & HotpotQA & 2Wiki & Musique & Bamboogle & M-Avg. & Avg. \\
      \midrule
      \rowcolor[gray]{0.9}
      Single + Multi & \textbf{0.470} & \textbf{0.615} & \textbf{0.444} & \textbf{0.510} & \textbf{0.420} & \textbf{0.412} & \textbf{0.183} & \textbf{0.371} & \textbf{0.371} & \textbf{0.416} \\
      Multi & 0.384 & 0.594 & 0.385 & 0.454 & \underline{0.395} & \underline{0.345} & \underline{0.155} & \underline{0.361} & \underline{0.314} & \underline{0.374} \\
      Single & \underline{0.459} & \underline{0.601} & \underline{0.438} & \underline{0.499} & 0.306 & 0.206 & 0.080 & 0.197 & 0.197 & 0.327 \\
      \bottomrule
    \end{tabular}
    }
\end{table} 

\subsection{Impact of Different Reward Modeling Methods}
\label{append:reward-design}

As described in Section~\ref{sec:reward modeling}, we adopt a conditional aggregation scheme for answer reward $r_\text{ans}$ and step-specific rewards $r_\text{query}$ \& $r_\text{merge}$, which are calculated over all the \token{search} and \token{merge} blocks in the rollout. To investigate the effect of reward design, we conduct additional experiments with three variants: (1) Unconditional: $r_\text{query}$ and $r_\text{merge}$ are applied regardless of whether the final answer is correct. (2) Turn-Level: $r_\text{query}$ and $r_\text{merge}$ are calculated at each turn, and the final reward is averaged over all turns. (3) EM: $r_\text{ans}$ is calculated using Extract Match, instead of F1 score. Results in Table~\ref{fig:abla-reward-design} show that our reward modeling method achieves higher overall accuracy.
This suggests that associating process rewards with the final outcome may help better align intermediate behaviors with task success, and that F1-based $r_\text{ans}$ could provide more informative feedback than EM in this setting.

\begin{table}[htbp]
    \caption{Comparison between different Reward Designs.}
    \label{fig:abla-reward-design}
    \centering
    \resizebox{\columnwidth}{!}{ 
    \begin{tabular}{lllllllll}
      \toprule
      Variants & NQ & TriviaQA & PopQA & HotpotQA & 2Wiki & Musique & Bamboogle & Avg. \\
      \midrule
      Our Method & 0.470 & 0.615 & 0.444 & 0.420 & 0.412 & 0.183 & 0.371 & 0.416 \\
      \midrule
      \rowcolor[gray]{0.9}
      \multicolumn{9}{c}{Conditional vs. Unconditional} \\
      Unconditional & 0.427 & 0.593 & 0.418 & 0.352 & 0.343 & 0.113 & 0.303 & 0.364 \\
      \hdashline
      \rowcolor[gray]{0.9}
      \multicolumn{9}{c}{Rollout-Level vs. Turn-Level} \\
      Turn-Level & 0.400 & 0.602 & 0.396 & 0.399 & 0.411 & 0.150 & 0.330 & 0.384 \\
      \hdashline
      \rowcolor[gray]{0.9}
      \multicolumn{9}{c}{F1 vs. EM} \\
      EM & 0.461 & 0.610 & 0.433 & 0.405 & 0.403 & 0.165 & 0.341 & 0.403 \\
      \bottomrule
    \end{tabular}
    }
\end{table}

\subsection{Impact of Different Merging Modules}
\label{append:merge-modules}

As described in Section~\ref{sec:multi-query}, the merging step is carried out by the search agent itself. We further analyze the effect of different merging strategies, including: (1) selecting the most relevant passages using TF-IDF (TF-IDF Screening), and (2) using an external LLM to merge retrieved information (LLM Extractor). To control for model size, we employ Qwen2.5-3B-Instruct as the external extractor. As shown in  Table~\ref{tab:abla-merge}, TF-IDF-based selection yields relatively poor results, suggesting that textual similarity to the question does not necessarily indicate usefulness.
The variants using an external 3B-LLM also achieve lower ultimate performance than MultiSearch, likely because the merging step is not integrated into the model’s optimization process. These observations highlight the potential benefits of our agent-driven merging paradigm.

\begin{table}[htbp]
    \caption{Comparison between different merging modules.}
    \label{tab:abla-merge}
    \centering
    \resizebox{\columnwidth}{!}{ 
    \begin{tabular}{lllllllll}
      \toprule
      Variants & NQ & TriviaQA & PopQA & HotpotQA & 2Wiki & Musique & Bamboogle & Avg. \\
      \midrule
      \rowcolor[gray]{0.9}
      MultiSearch & 0.470 & 0.615 & 0.444 & 0.420 & 0.412 & 0.183 & 0.371 & 0.416 \\
      w/ TF-IDF Screening & 0.383 & 0.517 & 0.375 & 0.286 & 0.244 & 0.116 & 0.218 & 0.308 \\
      w/ LLM Extractor & 0.407 & 0.577 & 0.397 & 0.341 & 0.341 & 0.136 & 0.288 & 0.355 \\
      \bottomrule
    \end{tabular}
    }
\end{table} 	 	 	 	 	 	 	 

\subsection{Comparison Between Base \& Instruct models}
\label{append:it-base}
\begin{figure}[htbp]
    \centering
    \includegraphics[width=\linewidth]{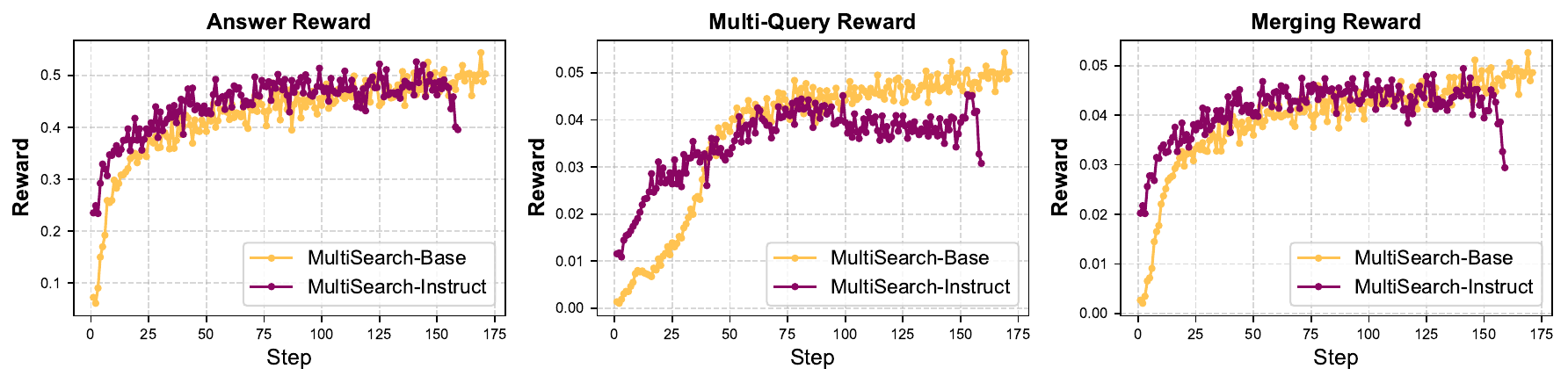}
    \caption{Training dynamics of base and instruct models.}
    \label{fig:append-it-base}
\end{figure}

We analyze the performance differences between the base and instruction-tuned models. As shown in Figure~\ref{fig:append-it-base}, the instruction-tuned model outperforms the base model in the early stages, but the base model gradually catches up, ultimately reaching comparable or even superior performance. This suggests that while instruction-tuning improves the model’s ability to follow human instructions, its prior knowledge may limit its generalization to multi-step reasoning tasks.

\subsection{Impact of Different Retrievers}
\label{append:retriever}
We compare two retriever variants: E5-base-v2 and BM25, with results reported in Table~\ref{tab:append-retriever}. As shown, E5-base-v2 achieves a better overall performance, suggesting that dense retrieval may provide more relevant evidence than lexical matching in our setting.
\begin{table}[htbp]
    \caption{Comparison between different retrievers.}
    \label{tab:append-retriever}
    \centering
    \resizebox{\columnwidth}{!}{ 
    \begin{tabular}{lllllllll}
      \toprule
      Variants & NQ & TriviaQA & PopQA & HotpotQA & 2Wiki & Musique & Bamboogle & Avg. \\
      \midrule
      E5 & 0.470 & 0.615 & 0.444 & 0.420 & 0.412 & 0.183 & 0.371 & 0.416 \\
      BM25 & 0.336 & 0.592 & 0.329 & 0.384 & 0.391 & 0.108 & 0.168 & 0.330 \\
      \bottomrule
    \end{tabular}
    }
\end{table}

\section{Prompts}
\label{append:prompts}

\subsection{Training Prompt for Search Agents}
\begin{tcolorbox}[colback=black!3!white, colframe=black!70!white, title=Search Agent Prompt]
\small
\textbf{Role.} You are a helpful assistant. Answer the given question with multi-turn search engine calling. You can search as many times as necessary.

\textbf{Input.} Question: \texttt{\{question\}}.

\textbf{Instructions.} Reason through the available information using \token{think} and\token{/think}.
Issue a search request using \token{search} $q_1, q_2, ...q_n$ \token{/search} when missing knowledge. The retrieved  documents will be placed in \token{information} and \token{/information}.
Generate three diverse search queries each time, applying one or more of these strategies: rephrasing, concept expansion, and question decomposition. 
Extract and integrate key information from the retrieved documents in \token{merge} and \token{/merge} after each search.

\textbf{Output.} Return a concise final answer inside \token{answer} and \token{/answer}, without detailed illustrations.
\end{tcolorbox}

\subsection{Prompt for LLM Extractor}
\begin{tcolorbox}[colback=black!3!white, colframe=black!70!white, title=LLM Extractor Prompt]
\small
\textbf{Role.} You are a document integration assistant.

\textbf{Input.} Question: \texttt{\{question\}}. Documents: \texttt{\{documents\}}. 

\textbf{Instructions.} Given a question and some documents, extract and integrate key information that is useful for the question. 

\textbf{Output.} Output only the integrated summary, without any extra commentary.
\end{tcolorbox}

\subsection{Prompt for SNR Evaluation}
\begin{tcolorbox}[colback=black!3!white, colframe=black!70!white, title=SNR Evaluation Prompt]
\small
\textbf{Role.} You are an evaluator of information quality. Your task is to assess the Signal-to-Noise Ratio of a given piece of content, based on a question and supporting documents. 

\textbf{Input.} Question: \texttt{\{question\}}. Supporting Documents: \texttt{\{documents\}}. Content to be evaluated: \texttt{\{content\}}. 

\textbf{Instructions.} "Signal" represents information that is directly relevant to answering the question. "Noise" represents information that is irrelevant to the question, redundant, contradictory to the supporting documents, or hallucinated. Read the question and the supporting documents carefully, analyze the content, and determine which parts are "signal" (useful, relevant, supported) and which parts are "noise" (useless, irrelevant, unsupported, or repetitive). 

\textbf{Output.} Provide a SNR between 0 and 1.
\end{tcolorbox}

\section{More Experimental Details}
\label{append:train-detail}

\subsection{Training Details}

MultiSearch is trained using the AdamW optimizer with a learning rate of  $1\times10^{-6}$  and a constant learning rate schedule without warmup. Weight decay is set to 0.01, and gradients are clipped with a maximum norm of 1.0. Training is conducted for 200 optimization steps using Fully Sharded Data Parallel (FSDP) without offloading parameters, gradients, or optimizer states. We use 4 NVIDIA A100 GPUs for 3B models, while larger configurations are trained on 8 NVIDIA A100 GPUs.

During training, rollout generation is performed using vLLM\footnote{\url{https://github.com/vllm-project/vllm}} with a sampling temperature of 1.0 and top-p set to 0.95. We adopt GDPO for policy optimization with a group size of 5. At each interaction step, the agent can issue up to 3 search queries, and the retrieval system returns the top 3 documents for each query. Key hyperparameters are summarized in Table~\ref{tab:append-hparams}.

% Increasing the number of queries or retrieved documents per step may result in longer rollouts due to more reasoning. However, the increase in generation time is relatively moderate. Given the observed performance gains shown in Section~\ref{sec:para-sensi}, we consider this trade-off acceptable for the present experiments.

Increasing the number of queries or retrieved documents per step may lead to longer rollouts due to the additional reasoning required. However, the resulting increase in generation time is relatively moderate. Moreover, because MultiSearch requires fewer reasoning steps, as shown in Figure~\ref{fig:avg-search-col}, the overall runtime remains comparable despite the longer per-step reasoning time. Given the performance gains observed in Section~\ref{sec:para-sensi}, we consider this trade-off acceptable for the present experiments.

\begin{table}[ht]
\centering
\caption{Hyperparameters used by MultiSearch.}
\label{tab:append-hparams}
\begin{tabular}{ll}
\toprule
\textbf{Hyper‑parameters} & \textbf{Value} \\ 
\midrule
Learning Rate      & $1\times10^{-6}$ \\
Training Batch Size      & 512  \\
Validation Batch Size    & 512  \\
Max Response Length     & 2048 \\
Micro Training Batch Size      & 32  \\
Group Size $G$ & 5\\
KL Coefficient $\beta$ & 0.001 \\
Max Search Turns     & 4 \\
Clip Ratio $\epsilon$ & 0.2 \\
Total Training Steps & 300 \\
\bottomrule
\end{tabular}
\end{table}

\subsection{Dataset Details}
The datasets we used are from FlashRAG Datasets\footnote{\url{https://huggingface.co/datasets/RUC-NLPIR/FlashRAG_datasets}}. Here, we present the detailed statistics in Table~\ref{tab:append-data}.

\begin{table}[ht]
\centering
\small
\caption{Statistics of the seven datasets.}
\label{tab:append-data}
\begin{tabular}{llllllll}
    \toprule
        \multirow{2}{*}{Split} & \multicolumn{3}{c}{Single-Hop QA} & \multicolumn{4}{c}{Multi-Hop QA} \\
      \cmidrule{2-8}
         & NQ & TriviaQA & PopQA & HotpotQA & 2Wiki & Musique & Bamboogle \\
        \midrule
        \# Train & 79168 & 78785 & -     & 90447 & 15,000 & 19,938 & -   \\
        \# Dev   & 8757  & 8837  & -     & 7405  & 12576  & 2417   & -   \\
        \# Test  & 3610  & 11313 & 14267 & -     & -      & -      & 125 \\
\bottomrule
\end{tabular}
\end{table}

\section{Case Studies}
Here, we present a case study comparing Search-R1 and MultiSearch in Table \ref{tab:placeholder}. 
The example illustrates that MultiSearch retrieves more relevant information through multi-query retrieval and further improves the SNR of the intermediate context through explicit merging.
\sethlcolor{yellow!50!white}
\begin{table*}[b]
    \centering
    \caption{Comparison between Search-R1 and  MultiSearch. The predictions are colored \textcolor{Red}{\textbf{Red}} for incorrect and \textcolor{Green}{\textbf{Green}} for correct answers. \hl{\textbf{Yellow segments}} denote the core evidences.}
    \begin{tabular}{p{0.98\linewidth}}
        \toprule
         \textbf{Question 1}: What is the name of the Sunday before Easter Sunday? \\
         \midrule
         Ground Truth: Palm Sunday. \\
         \midrule
         \rowcolor[gray]{0.95}
         \textbf{Search-R1} \\
         \# \textcolor{violet}{\texttt{Think:}} Easter Sunday is a Christian holiday celebrating the resurrection of Jesus Christ. Sunday before Easter Sunday falls on a different name due to its significance in the Christian calendar. \\
         \# \textcolor{purple}{\texttt{Search:}} What is the name of the Sunday before Easter Sunday? \\
         \# \textcolor{orange!95!black}{Info:} ... Doc 2(Title: Easter) ...The week beginning with Easter Sunday ... \hl{The day before Easter is properly called Holy Saturday}. Doc 3(Title: Quinquagesima) ... is one of the names used in the Western Church for the Sunday before Ash Wednesday...  the fifty days before Easter Day using inclusive counting which counts both Sundays ... \\
         \# \textcolor{gray}{Answer:} \textcolor{Red}{\textbf{Good Friday.}} \\
         \midrule
         \rowcolor[gray]{0.95}
         \textbf{MultiSearch} \\
         \# \textcolor{violet}{\texttt{Think:}} First, we need to find out what is the name of the Sunday that comes before Easter Sunday. We can search directly or by asking for the name of the week before Easter Sunday. \\
         \# \textcolor{purple}{\texttt{Search:}} \textbf{name of the sunday before easter sunday}, \textbf{name of the week before easter sunday}, \textbf{when is the sunday before easter sunday} \\
         \# \textcolor{orange!95!black}{\texttt{Info:}} Doc 1(Title: Easter) ... \hl{The Sunday before Easer is Palm Sunday}, with the Wednesday before Easter being known as Spy Wednesday ... Doc 5(Title: "Holy Week") Holy Week in Christianity is \hl{the week just before Easter} ... the last week of Lent, in the West, – \hl{Palm Sunday}, Holy Wednesday (Spy Wednesday), Maundy Thursday (Holy Thursday), ... – are all included. \\
         \# \textcolor{brown}{\textbf{Merge:}} From the retrieved documents, we can see that the Sunday before Easter Sunday is Palm Sunday. \\
         \# \textcolor{gray}{Answer:} \textcolor{Green}{\textbf{Palm Sunday.}} \\
         \bottomrule
         \toprule
         \textbf{Question 2}: Are Portland International Airport and Gerald R. Ford International Airport both located in Oregon? \\
         \midrule
         Ground Truth: No. \\
         \midrule
         \rowcolor[gray]{0.95}
         \textbf{Search-R1} \\
         \# \textcolor{violet}{\texttt{Think:}} To answer this question, I need to determine if both Portland International Airport and Gerald R. Ford International Airport are located in the same geographic location, specifically Oregon. \\
         \# \textcolor{purple}{\texttt{Search:}} Are Portland International Airport and Gerald R. Ford International Airport both located in Oregon?  \\
         \# \textcolor{orange!95!black}{Info:} Doc 1(Title: "Portland International Airport") \hl{Portland International Airport} is ... and \hl{the largest airport in the U.S. state of Oregon} that accounts for ... \\
         \# \textcolor{violet}{\texttt{Think:}} From the information provided, I can conclude that both Portland International Airport and Gerald R. Ford International Airport are located in Oregon.  \\
         \# \textcolor{gray}{Answer:} \textcolor{Red}{\textbf{Yes.}} \\
         \midrule
         \rowcolor[gray]{0.95}
         \textbf{MultiSearch} \\
         \# \textcolor{violet}{\texttt{Think:}} First, we need to determine the locations of Portland International Airport and Gerald R. Ford International Airport. Then, we'll check if both airports are in Oregon. \\
         \# \textcolor{purple}{\texttt{Search:}} \textbf{where is Portland International Airport}, \textbf{where is airport Gerald R. Ford International Airport}, \textbf{are Portland International Airport located in Oregon} \\
         \# \textcolor{orange!95!black}{\texttt{Info:}} Doc 1(Title: "Portland International Airport") \hl{Portland International Airport} is ... and \hl{the largest airport in the U.S. state of Oregon} that accounts for ...
         Doc 4(Title: "Gerald R. Ford International Airport") ... \hl{Gerald R. Ford International Airport} is a commercial airport in Cascade Township approximately \hl{southeast of Grand Rapids, Michigan} ... \\
         \# \textcolor{brown}{\textbf{Merge:}} From the retrieved documents, we can see that \hl{Portland International Airport is in Oregon and Gerald R. Ford International Airport is in Michigan}. The two airports are not in the same state. \\
         \# \textcolor{gray}{Answer:} \textcolor{ForestGreen}{\textbf{No.}} \\
         \bottomrule
    \end{tabular}
    \label{tab:placeholder}
\end{table*}

\section{Limitations}
\label{sec:limitations}
Despite the promising results, our work has several limitations:
\paragraph{Limited Task Generalization.}  
The experiments are conducted exclusively on question-answering datasets following prior work, without other types of tasks or domains. As a result, the generality of MultiSearch beyond QA settings remains to be explored.

\paragraph{Predefined Retrieval Corpus.}  
The method relies on a static retrieval corpus for all experiments. Its performance in fully online search environments, such as live web search engines, has not been tested, which may limit its applicability in dynamic information retrieval scenarios.

\paragraph{Fixed Query Generation Strategies.}  
While we explore multiple strategies for generating queries, the underlying framework still operates within a relatively fixed set of strategies. This may restrict flexibility in more open-ended or unstructured reasoning tasks.

\section{Broader Impacts}

MultiSearch explores a parallel search and explicit merging framework. By encouraging the agent to maintain multiple candidate searching directions, it can help broaden the scope of information acquisition during the reasoning process, particularly in scenarios where relevant evidence is distributed across multiple sources. This suggests a possible direction for dynamically balancing search breadth and reasoning depth depending on task difficulty.
Furthermore, by making the merging process explicit, MultiSearch introduces a more structured interface between retrieval and reasoning. This “explore-then-merge” pattern encourages the agent to organize retrieved evidence prior to final inference, contributing to a clearer understanding of how retrieved passages are integrated. In this sense, the framework could facilitate more detailed post-hoc analysis of evidence usage, although we do not explicitly evaluate interpretability in this work. 

% \clearpage
% \input{checklist}
	 	 	 	 	 	 	 	
\end{document}